%% file: main.tex
\documentclass[10pt,twocolumn,letterpaper]{article}

\usepackage[pagenumbers]{cvpr} 

\input{preamble}

\definecolor{cvprblue}{rgb}{0.21,0.49,0.74}
\usepackage[pagebackref,breaklinks,colorlinks,citecolor=cvprblue]{hyperref}

\def\paperID{36} 
\def\confName{ECV}
\def\confYear{2024}

\title{Robust Analysis of Multi-Task Learning Efficiency: New Benchmarks on Light-Weighed Backbones and Effective Measurement of Multi-Task Learning Challenges by Feature Disentanglement}

\author
{
    Dayou Mao$^{1}$\\{\tt\small daniel.mao@uwaterloo.ca}
    \and
    Yuhao Chen$^{1}$\\{\tt\small yuhao.chen1@uwaterloo.ca}
    \and
    Yifan Wu$^{1}$\\{\tt\small yifan.wu1@uwaterloo.ca}
    \and
    Maximilian Gilles$^{2}$\\{\tt\small maximilian.gilles@kit.edu}
    \and
    Alexander Wong$^{1}$\\{\tt\small a28wong@uwaterloo.ca}
    \and
    $^{1}$Vision and Image Processing Research Group, Waterloo, Canada\\
    $^{2}$Karlsruhe Institute of Technology (KIT), Karlsruhe, Germany
}

\begin{document}
\maketitle
\input{sections/0_abstract}    
\input{sections/1_intro}
\input{sections/2_related_work}

\input{sections/3_benchmarks}

\input{sections/4_surrogate}
\input{sections/5_method}

\input{sections/6_conclusions}
{
    \small
    \bibliographystyle{ieeenat_fullname}
    \bibliography{main}
}
\input{sections/suppl}

\end{document}

%% file: preamble.tex
\usepackage[dvipsnames]{xcolor}

\usepackage{array}
\newcolumntype{P}[1]{>{\centering\arraybackslash}p{#1}}

%% file: sections/0_abstract.tex
\begin{abstract}
    One of the main motivations of MTL is to develop neural networks capable of inferring multiple tasks simultaneously. While countless methods have been proposed in the past decade investigating robust model architectures and efficient training algorithms, there is still lack of understanding of these methods when applied on smaller feature extraction backbones, the generalizability of the commonly used fast approximation technique of replacing parameter-level gradients with feature level gradients, and lack of comprehensive understanding of MTL challenges and how one can efficiently and effectively identify the challenges. In this paper, we focus on the aforementioned efficiency aspects of existing MTL methods. We first carry out large-scale experiments of the methods with smaller backbones and on a the MetaGraspNet dataset as a new test ground. We also compare the existing methods with and without using the fast gradient surrogate and empirically study the generalizability of this technique. Lastly, we propose Feature Disentanglement measure as a novel and efficient identifier of the challenges in MTL, and propose Ranking Similarity score as an evaluation metric for different identifiers to prove the faithfulness of our method.
\end{abstract}

%% file: sections/1_intro.tex
\section{Introduction}\label{sec:intro}

\begin{figure}[t]
    \centering
    \includegraphics[width=\linewidth]{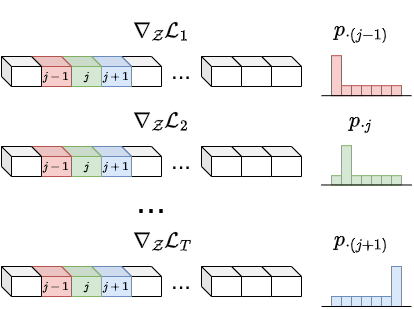}
    \caption
    {
        Illustration of feature disentanglement calculation. In the above, $p_{\cdot j}$ denotes the mapping $i \mapsto p_{ij}$, which is the (smoothened) distribution of feature saliency at location $j$ across all tasks. Same for $p_{\cdot(j-1)}$ and $p_{\cdot(j+1)}$. If an extracted feature is disentangled for the $T$ down stream tasks, then each distribution $p_{\cdot j}$ should be concentrated on fewer tasks and have lower entropy.
    }
    \label{fig:def_fd}
\end{figure}


Multi-task learning (MTL) is a learning paradigm that aims to develop systems capable of performing multiple tasks simultaneously.
At a high-level, MTL designs systems with two key components. They learn a single backbone that encodes raw inputs into shared representations and append prediction heads that consume the shared representation and each learning a different down stream task atop the backbone.
Multi-task models are efficient because the majority of parameters lie in the feature extraction backbone and are \textit{shared parameters}, while the \textit{task-specific parameters} are usually very light-weighed.
The merit of designing MTL systems is believed to be two-fold. Firstly, compared to developing a single-task model for each task, multi-task models significantly reduces overall model size because of shared backbone, and benefits of faster inference speed, reduced memory footprint, and lower power consumption all follow. Secondly, it is a common belief that the more complex the supervision signals, the richer the learned representation \cite{liu2019self, achituve2021self, navon2020auxiliary}. Introducing supervision signals from a diverse range of down stream tasks has been proven to be an effective approach to improve performance compared to training single-task learning systems \cite{ruder2017overview, zamir2018taskonomy, caruana1997multitask, neven2017fast, blitznet, padnet, mgda, uncertainty, pinto2017learning}. The idea of MTL has already been exploited in classical discriminative computer vision algorithms well before it becomes an active research topic on its own \cite{overfeat, fastrcnn, maskrcnn}.


Despite all the potential benefits MTL can bring, training a multi-task models is also more challenging, as observed in the fields of
computer vision \cite{maskrcnn, uncertainty, cosreg, ubernet, zhao2018modulation, standley2020tasks},
natural language processing \cite{wang2020negative, wang2019characterizing},
meta learning \cite{abdollahzadeh2021revisit},
and reinforcement learning \cite{parisotto2015actor, rusu2015policy, teh2017distral, hessel2019multi, schaul2019ray}.
Challenges of MTL arise because it requires consideration of multiple objectives when training the single shared backbone. Given vision tasks with a wide range of difficulties, output dimensions, and types of training loss functions, it is rarely the case that all tasks ``align well'' during training. Namely, a parameter update on the backbone that improves performance of one task may lead to worse performance of another task at the same time. This phenomenon is known as \textit{negative transfer} or \textit{destructive interference} \cite{zhao2018modulation}.
It is widely accepted that this phenomenon can be explained by two factors: \textit{task conflicts} and \textit{task dominance} (Section \ref{subsec:preliminaries}).


Extensive research has delved into MTL techniques aimed at mitigating the issues of task conflicts and task dominance, as we will elaborate upon in more detail in Section \ref{sec:related_work}. However, the literature still lacks understanding in the following 4 aspects.
Firstly, the literature has been focusing on ResNet50 \cite{mgda, imtl, alignedmtl} and SegNet \cite{pcgrad, cagrad, randomweighting, nashmtl, alignedmtl} backbones, but lack attention to smaller backbones like ResNet18. 
Secondly, the existing literature is mostly focused on
toy examples \cite{pcgrad, graddrop, cagrad, nashmtl, alignedmtl, gradnorm, pareto, famo, li2017self},
Celeb-A \cite{pcgrad, graddrop, randomweighting, mgda, imtl, famo},
CityScapes \cite{pcgrad, randomweighting, mgda, cagrad, nashmtl, alignedmtl, uncertainty, imtl, famo},
NYU/NYU-v2 \cite{pcgrad, randomweighting, cagrad, nashmtl, alignedmtl, gradnorm, imtl, yun2023achievement, famo, cosreg},
Multi-MNIST \cite{pcgrad, mgda, kumar2010self, cosreg},
Multi-Fashion-MNIST \cite{cagrad, pareto},
Multi-Task CIFAR-100 \cite{pcgrad, graddrop},
and QM-9 \cite{nashmtl, famo}, but lack common test ground on complex vision tasks across diverse range of tasks with a large-scale dataset.
Thirdly, prior work \cite{mgda, alignedmtl, graddrop, rotograd} have proposed the technique to use feature-level gradients to replace parameter-level gradients for more efficient computation, which we call the ``Fast Gradient Surrogate'' technique. However, some \cite{rotograd} lack theoretical guarantee or solid empirical analysis to prove effectiveness of this surrogate and others \cite{nashmtl} have reported poor results when doing so. A solid understanding of its generalizability is still missing.
Fourthly, prior work (Sections \ref{subsec:related_manipulation}, \ref{subsec:related_balancing}, \ref{subsec:related_regularization}) hypothesize that task conflicts and task dominance are the root causes for poor performance of MTL models. But there is lack of analysis on the causal relationship between these claimed root cause and model performance.
Moreover, both of these measures require computing the gradients of all losses w.r.t. shared parameters and hence requires back propagation through the entire backbone $T$ times if $T$ is the number of tasks, which is super computationally expensive \cite{mgda, rotograd, alignedmtl}. A more efficient way to understand MTL challenges is needed.

In this paper, we conduct large-scale experiments to address the efficiency issues of optimization algorithms for MTL from the following angles: We provide comprehensive benchmark results with ResNet18 \cite{resnet} backbones on MetaGraspNet \cite{metagraspnetv2}, CityScapes \cite{cityscapes}, and NYU-v2 \cite{nyu_v2} datasets, compare existing methods with their fast gradient surrogates, and propose Feature Disentanglement measure, shown in Figure \ref{fig:def_fd}, as a novel perspective to identify MTL challenges and propose Ranking Similarity as an evaluation protocol for comparing different identifiers. Our contribution in this paper is three-fold:
\begin{itemize}
    \item We complement the existing performance benchmarks in the literature with smaller feature extraction backbones, ResNet18 \cite{resnet}, and a more robust benchmark dataset, MetaGraspNet \cite{metagraspnetv2}. (Section \ref{sec:benchmarks}).
    \item We provide solid empirical evidence and show that the fast gradient surrogate technique cannot not be generalized to all methods and has to be analyzed and tested depending on the specific algorithm (Section \ref{sec:surrogate});
    \item We propose Feature Disentanglement measurement that can more efficiently and faithfully reflect MTL challenge compared to traditional task conflicts and task dominance measures, evaluated by Ranking Similarity against test-time performance. (Section \ref{sec:method}).
\end{itemize}

%% file: sections/2_related_work.tex
\section{Problem Definition and Related Work}\label{sec:related_work}

The problem of Multi-Task Learning aims to train a model to learn several tasks simultaneously. Formally, we assume for the following notations.
$T$ denotes the number of tasks.
$\mathcal{X}$ denotes the set of training inputs.
$\mathcal{Y}_{i}$ denotes the label space for task $i$ and we define $\mathcal{Y} := \bigoplus_{i=1}^{T}\mathcal{Y}_{i}$ as the collection of all task labels.
$\mathcal{D} \subset \mathcal{X} \times \mathcal{Y}$ denotes the training dataset.
$f_{\theta}: \mathcal{X} \to \mathcal{Y}$ denotes a neural network parametrized by $\theta \in \Omega$ where $\Omega$ is the parameter space.
$\mathcal{L}_{i}: \mathcal{Y}_{i} \times \mathcal{Y}_{i} \to \mathbb{R}_{\geq0}$ denotes the loss functions for the $i$-th task.
In general, multi-task learning can be formulated as the following optimization problem:
\begin{equation}
    \displaystyle
    \min_{\theta \in \Omega}
    \sum_{(x, y) \in \mathcal{D}}\mathcal{L}(f_{\theta}(x), y),
    \label{eq:formulation}
\end{equation}
where $\mathcal{L}$ is some loss function, either scalar-valued or vector-valued, defined on $\mathcal{Y} \times \mathcal{Y}$.
The average of all task losses $\mathcal{L} := \frac{1}{T}\sum_{i=1}^{T}\mathcal{L}_{i} \in \mathbb{R}_{\geq0}$, or Equal Weighting (EW), is the most commonly used baseline. Pareto optimization minimizes the loss vector $\mathcal{L} := (\mathcal{L}_{1}, ..., \mathcal{L}_{T})^{\top} \in \mathbb{R}_{\geq0}^{T}$ w.r.t. the partial-order $\leq_{K}$ on $\mathbb{R}^{T}$ induced by the pointed, closed, and convex cone $K := \mathbb{R}_{\geq0}^{T}$ \cite{mgda, pareto, nashmtl, ye2021multi}.
As we discuss in the remaining of this section, wealth of research has been done on efficient and effective optimization algorithms that solve Problem \ref{eq:formulation} and on designing the specific forms of the loss function $\mathcal{L}$.

We make two remarks here on the differences and relationships between gradient manipulation methods (Section \ref{subsec:related_manipulation}) and gradient balancing methods (Section \ref{subsec:related_balancing}), before we go into depth. (1) Due to the linearity of the gradient operator $\nabla$, scaling the loss function and scaling the gradients are essentially the same. However, gradient manipulation methods aim to manipulate the \textit{directions} of the gradients to resolve \textit{task conflicts}, whereas gradient balancing methods aim to manipulate the \textit{magnitudes} of the gradients to resolve \textit{task dominance}. (2) Gradient manipulation methods focus on manipulating gradients for the shared parameters and tune the task-specific parameters as usual single-task learning, whereas gradient balancing methods scales gradients for all model parameters.

\textbf{Additional Notations and Terminologies}.
Throughout, we adopt the following notations: (1) for any natural number $n$, $[n] := \{1, ..., n\}$, (2) $\theta^{\text{sh}} \in \mathbb{R}^{d}$ denotes shared parameters where $d$ is the number of shared parameters and $\mathcal{Z}$ denotes shared representation, (3) $g_{i} := \nabla_{\theta^{\text{sh}}}\mathcal{L}_{i}$ denotes each task gradient, or ``parameter-level gradients'', and $G := [g_{1}, ..., g_{T}]^{\top}$ denotes the gradient matrix whose columns are the task gradients, (4) $\nabla_{\mathcal{Z}}\mathcal{L}_{i}$ are the ``feature-level gradients''.

\input{sections/2_related_work/1_gradient_manipulation}
\input{sections/2_related_work/2_gradient_balancing}
\input{sections/2_related_work/3_gradient_regularization}
\input{sections/2_related_work/4_surrogate}
\input{sections/2_related_work/5_xai}

%% file: sections/2_related_work/1_gradient_manipulation.tex
\subsection{Gradient Manipulation}\label{subsec:related_manipulation}

To update the shared parameters taking all tasks into consideration, gradient manipulation methods 
\cite{randomweighting, pcgrad, gradvac, graddrop,
mgda, cagrad, itmtl, nashmtl, alignedmtl}
compute task gradients $g_{i} := \nabla_{\theta^{\text{sh}}}\mathcal{L}_{i}$ and each propose different method to compute $\alpha = (\alpha_{1}, ..., \alpha_{T})^{\top} \in \mathbb{R}^{T}$ so that the final update on the shared parameters is done by the aggregation $\hat{g} := \sum_{i=1}^{T}\alpha_{i}g_{i}$. Prediction heads are trained as in usual single-task learning, each supervised by their own loss function. Computation of $\alpha$ can be done either explicitly or implicitly.

\textbf{Explicit Methods} \cite{randomweighting, pcgrad, gradvac, graddrop} derive closed-form formulae for $\alpha$ based on some heuristics and may rely on some stochasticity. PCGrad \cite{pcgrad} proposed to reduce task conflicts by projecting gradients onto the normal planes of each other, whenever the angle between two gradient vectors is obtuse. GradVac \cite{gradvac} expanded along this direction and proposed to encourage acute angles between gradients by maintaining an Exponential Moving Average (EMA) of cosine similarity between task gradients as upper bound on angles between. GradDrop \cite{graddrop} propose to mask the gradients with Gradient Sign Purity so that the gradient signs are more aligned. Random Gradient Weighting (RGW) \cite{randomweighting} draws random samples from a Gaussian distribution, normalize them into a probability simplex, and re-weigh the task gradients with normalized samples.

\textbf{Implicit Methods} \cite{mgda, cagrad, itmtl, nashmtl, alignedmtl} hypothesize different objectives and compute the gradient weights $\alpha$ by solving either an optimization problem or a system of equations. MGDA \cite{mgda} re-weighs task gradients so that the result is the norm minimizer in the convex hull enclosed by the task gradients. CAGrad \cite{cagrad} proposed to maximize the minimum amount of decrease (in absolute value) in the individual losses and could be viewed as a generalized version of MGDA by adding a search region. Nash-MTL \cite{nashmtl} follows a similar idea as CAGrad, but rather than maximizing the minimum amount of decrease, it maximizes the sum of the log decreases in each individual loss. This eventually resolves to solving a (non-linear) system of equations. Aligned-MTL \cite{alignedmtl} proposed to first approximate the gradient matrix $G := [g_{1}, ..., g_{T}] \in \mathbb{R}^{d \times T}$ with its best unitary matrix approximation, which is sure to have stability number $1$, and then use the summation of the approximated task gradients to update $\theta^{\text{sh}}$.

%% file: sections/2_related_work/2_gradient_balancing.tex
\subsection{Gradient Balancing}
\label{subsec:related_balancing}

Similar to gradient manipulation methods reviewed in the previous section, gradient balancing methods
\cite{randomweighting, gradnorm, prioritization, yun2023achievement, imtl, mtan,
pareto, famo,
uncertainty, jacob2023online,
rudd2016moon},
or loss balancing methods, aim to assign weights $w = (w_{1}, ..., w_{T})^{\top} \in \mathbb{R}_{\geq0}^{T}$ to the loss functions and solve the following optimization problem, so that all tasks are learned at compatible pace:
\begin{equation}
    \displaystyle
    \min_{\theta \in \Omega}
    \sum_{(x, y) \in \mathcal{D}}\sum_{i=1}^{T}w_{i}\mathcal{L}_{i}(f_{\theta}(x), y_{i})
    \label{eq:task}
\end{equation}
Computation of $w$ can be done either explicitly or implicitly, or even by an extra optimizer.

\textbf{Explicit Methods} \cite{randomweighting, gradnorm, prioritization, yun2023achievement, imtl, mtan} compute $w$ explicitly based on different heuristics. GradNorm \cite{gradnorm} proposed to control the learning pace of different tasks based on the relative norm of the gradients and does so by assigning task weights and updating them at each training iteration. Exploiting the same idea as RGW \cite{randomweighting}, the same paper also proposed Random Loss Weighting (RLW). In \cite{mtan}, the authors propose yet another simple weighting mechanism named Dynamic Weight Average (DWA) to re-weight the losses based on the relative descending rate of each loss.

\textbf{Implicit Methods} \cite{pareto, famo} compute the loss weights $w$ by solving an optimization subproblem at each training iteration. In particular, FAMO \cite{famo} proposed that the parameter update at each training step should maximize the lowest relative improvement of the task losses. The subtle difference from CAGrad \cite{cagrad} is that FAMO uses \textit{relative} improvements, so that solving the optimization subproblem results in reducing gradient dominance, whereas CAGrad uses \textit{absolute} improvements and would result in reduced gradient conflicts.

\textbf{Optimization-Based Methods} \cite{uncertainty, jacob2023online} dynamically update the loss weights with an extra set of objective and optimizer.

%% file: sections/2_related_work/3_gradient_regularization.tex
\subsection{Gradient Regularization}
\label{subsec:related_regularization}

Gradient regularization methods \cite{cosreg, rotograd} design regularization terms in the loss function and solves the following optimization problem:
\begin{equation}
    \displaystyle
    \min_{\theta \in \Omega}
    \sum_{(x, y) \in \mathcal{D}} \mathcal{L}(f_{\theta}(x), y) + \mathcal{L}_{\text{reg}}(G)
\end{equation}
where $G \in \mathbb{R}^{d \times T}$ is the gradient matrix.
In particular, Suteu \cite{cosreg} argues that orthogonal task gradients are beneficial to learning and adds squared cosine similarity as a regularization term. Javaloy \cite{rotograd} proposed to regularize the gradients by minimizing the angles between each task gradient and their average.

%% file: sections/2_related_work/4_surrogate.tex
\subsection{Fast Gradient Surrogate}

Most of the existing MTL algorithms rely on computing parameter-level gradients $\nabla_{\theta^{\text{sh}}}\mathcal{L}_{i}$, which requires back propagation through the entire backbone where the shared parameters lie.
Feature-level gradients $\nabla_{\mathcal{Z}}\mathcal{L}_{i}$ are a lot cheaper to compute as back propagation is only required through the prediction heads, which are usually very light-weighed.
However, some works \cite{mgda, alignedmtl} provide theoretical reasoning that the optimization problem could be solved with feature-level gradients $\nabla_{\mathcal{Z}}\mathcal{L}_{i}$, as they define an upper bound for the objective function. Other works \cite{rotograd, nashmtl} have also empirically investigated application of this feature-level surrogate but \cite{rotograd} obtained reasonable results while \cite{nashmtl} observed poor results.

%% file: sections/2_related_work/5_xai.tex
\subsection{Saliency Maps in Explainable AI}

Explainable AI seeks to explain behaviours of AI systems. The representative work, GradCam \cite{gradcam}, of top-down approaches proposed to compute a saliency map for the intermediate activations via the (absolute value of) gradients of class scores w.r.t. the activations. This has inspired us to quantify feature disentanglement by measuring the saliency for each task. More details are explained in Section \ref{sec:method}.

%% file: sections/3_benchmarks.tex
\section{Benchmarks}\label{sec:benchmarks}

In this section, we carry out large-scale experiments to complement the understanding of MTL optimization algorithms in the following two dimensions: (1) application on smaller models, namely ResNet18 backbones, and (2) application in efficiency demanding robotics vision application, on a more complex and super large-scale real-world dataset, the MetaGraspNet dataset \cite{metagraspnetv2}. We also provide results on CityScapes \cite{cityscapes} and NYU-v2 \cite{nyu_v2} for completeness.

\noindent\textbf{Selected Methods}. We studied 15 MTL optimization algorithms, from three categories: (1) we selected PCGrad \cite{pcgrad}, GradVac \cite{gradvac}, GradDrop \cite{graddrop}, RGW \cite{randomweighting}, MGDA \cite{mgda}, CAGrad \cite{cagrad}, Nash-MTL \cite{nashmtl}, and Aligned-MTL \cite{alignedmtl} from the gradient manipulation category (8 in total), (2) we selected Uncertainty \cite{uncertainty}, GradNorm \cite{gradnorm}, IMTL \cite{imtl}, FAMO \cite{famo}, RLW \cite{randomweighting}, and DWA \cite{mtan} from the gradient balancing category (6 in total), and (3) CosReg \cite{cosreg} from the gradient regularization category. Among these methods, MGDA and Aligned-MTL have provided theoretical analysis on replacing parameter-level gradients with feature-level gradients, known as MGDA-UB and Aligned-MTL-UB, respectively. These variants are also benchmarked. A comprehensive study on this fast approximation technique is reported in Section \ref{sec:surrogate}.

\noindent\textbf{General Setup}. For stochasticity consideration, all experiments were repeated 3 times and all results in this paper are average results across the 3 repetitions. As the focus of this paper is to study relative performance as compared to the baseline, rather than proposing novel MTL methods, we focus our experiments on the early stage of training. We refer the readers to Appendix \ref{suppl:setup_benchmarks} for more details.

\input{sections/3_benchmark/1_metagraspnet}
\input{sections/3_benchmark/2_city_scapes_nyu_v2}

%% file: sections/3_benchmark/1_metagraspnet.tex
\subsection{Experiments on MetaGraspNet Dataset}
\label{subsec:benchmark_metagraspnet}

\begin{figure}[t]
    \centering
    \includegraphics[width=\linewidth]{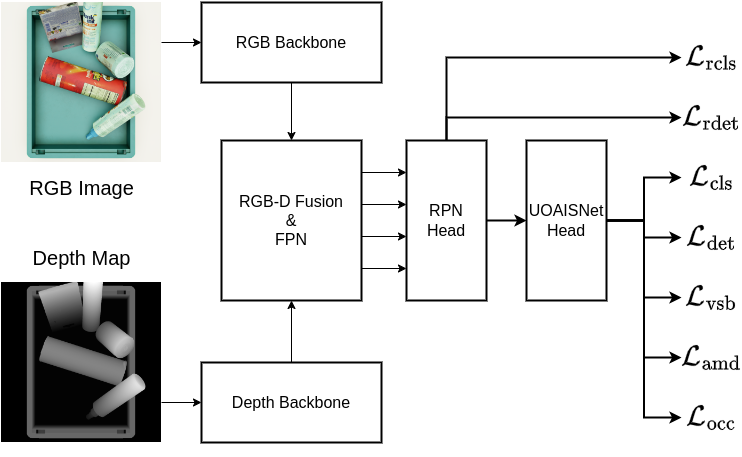}
    \caption{Illustration of model architecture used on the MetaGraspNet \cite{metagraspnetv2} benchmark.}
    \label{fig:network}
\end{figure}

\noindent\textbf{The Dataset}. We provide a new test ground for multi-task learning on the MetaGraspNet dataset \cite{metagraspnetv2}, due to its significantly larger dataset size, increased task complexity, and higher real-world value.

\noindent\textbf{Network Architecture}. We follow \cite{uoaisnet, fastgraspnext} and utilize two ResNet \cite{resnet} backbones, one for RGB image input and one for depth map input. Results at each stage of the ResNet from the RGB image input and the depth map input are fused by convolution layers. These fused features collectively yield the output from the backbone network. Then we feed the backbone outputs to a Feature Pyramid Network (FPN) \cite{fpn} to fuse the features from different levels. On top of this extracted feature from the FPN neck we attach Region Proposal Network (RPN) \cite{fasterrcnn} and UOAISNet \cite{uoaisnet} as prediction heads for amodal object bounding boxes, visible object masks, amodal object masks, and occlusion predictions. An architecture overview is shown in Figure \ref{fig:network}.

\noindent\textbf{Training Objective and Baseline Definition}. We follow \cite{fastgraspnext} and use the following loss functions: $\mathcal{L}_{\text{rcls}}$ and $\mathcal{L}_{\text{rdet}}$ for foreground/background classification and bounding boxes regression by the RPN head and $\mathcal{L}_{\text{cls}}$, and $\mathcal{L}_{\text{det}}$ by the amodal detection head; $\mathcal{L}_{\text{vsb}}$ for visible object masks prediction, $\mathcal{L}_{\text{amd}}$ for amodal object masks prediction, and $\mathcal{L}_{\text{occ}}$ for object occlusion prediction. We set the baseline loss function to be
\begin{equation}\label{eq:tot}
    \mathcal{L}_{\text{tot}} :=
    \mathcal{L}_{\text{rcls}} +
    \mathcal{L}_{\text{rdet}} +
    \mathcal{L}_{\text{cls}} +
    \mathcal{L}_{\text{det}} +
    \mathcal{L}_{\text{vsb}} +
    \mathcal{L}_{\text{amd}} + 
    \mathcal{L}_{\text{occ}},
\end{equation}
i.e., the sum of all 7 loss functions.

\noindent\textbf{Experiment Results}. Note that experiments are not meant to replicate existing results but rather comparing performance of different MTL optimization algorithms against the baseline. However, we observed extremely poor performance with GradNorm \cite{gradnorm} and Nash-MTL \cite{nashmtl}, and the feature-level gradient counterpart of GradDrop \cite{graddrop}, which is what originally proposed, so these methods are not reported on MetaGraspNet dataset. Results have shown that GradVac \cite{gradvac}, GradDrop \cite{graddrop}, IMTL \cite{imtl}, DWA \cite{mtan}, MGDA-UB \cite{mgda}, (rep) CAGrad \cite{cagrad}, and (rep) CosReg \cite{cosreg} achieved consistent performance gain compared to the baseline on all 6 evaluation metrics. GradVac \cite{gradvac}, IMTL \cite{imtl}, and (rep) MGDA achieved top-three performance under a majority ($\geq 3$) of the metrics. Full results on MetaGraspNet are summarized in Table \ref{table:benchmark_metagraspnet}.

\begin{table*}
    \centering
    \small{\input{tables/benchmark_metagraspnet}}
    \caption
    {
        Benchmark results of all selected methods with ResNet-18 backbone on MetaGraspNet \cite{metagraspnetv2} dataset. Performance increase (with \color{red}$\uparrow$\color{black}) or decrease (with \color{blue}$\downarrow$\color{black}) that's more than $0.01$ are shown in brackets after each table entry. Scores within $0.01$ offset from the baseline are treated as comparable performance and labeled by ``--''. Best viewed in color.
    }
    \label{table:benchmark_metagraspnet}
\end{table*}

%% file: tables/benchmark_metagraspnet.tex
\begin{tabular}{c|cccccc}
    \hline
     & BBox mAP & BBox mAR & VMask mAP & VMask mAR & AMask mAP & AMask mAR \\
    \hline
    Baseline & 0.383 & 0.519 & 0.518 & 0.647 & 0.490 & 0.617 \\
    \hline
    RGW \cite{randomweighting} & 0.358 (\color{blue}$\downarrow$\color{black}) & 0.513 (\color{blue}$\downarrow$\color{black}) & 0.490 (\color{blue}$\downarrow$\color{black}) & 0.644 (--) & 0.462 (\color{blue}$\downarrow$\color{black}) & 0.614 (--) \\
    PCGrad \cite{pcgrad} & 0.370 (\color{blue}$\downarrow$\color{black}) & 0.526 (\color{red}$\uparrow$\color{black}) & 0.500 (\color{blue}$\downarrow$\color{black}) & 0.657 (\color{red}$\uparrow$\color{black}) & 0.454 (\color{blue}$\downarrow$\color{black}) & 0.595 (\color{blue}$\downarrow$\color{black}) \\
    GradVac \cite{gradvac} & 0.393 (\color{red}$\uparrow$\color{black}) & \textbf{0.560} (\color{red}$\uparrow$\color{black}) & 0.521 (--) & \textbf{0.680} (\color{red}$\uparrow$\color{black}) & 0.495 (--) & \textbf{0.654} (\color{red}$\uparrow$\color{black}) \\
    MGDA \cite{mgda} & 0.371 (\color{blue}$\downarrow$\color{black}) & 0.531 (\color{red}$\uparrow$\color{black}) & 0.452 (\color{blue}$\downarrow$\color{black}) & 0.594 (\color{blue}$\downarrow$\color{black}) & 0.430 (\color{blue}$\downarrow$\color{black}) & 0.564 (\color{blue}$\downarrow$\color{black}) \\
    CAGrad \cite{cagrad} & \textbf{0.411} (\color{red}$\uparrow$\color{black}) & 0.557 (\color{red}$\uparrow$\color{black}) & 0.522 (--) & 0.648 (--) & 0.488 (--) & 0.604 (\color{blue}$\downarrow$\color{black}) \\
    GradDrop \cite{graddrop} & 0.399 (\color{red}$\uparrow$\color{black}) & 0.541 (\color{red}$\uparrow$\color{black}) & 0.533 (\color{red}$\uparrow$\color{black}) & 0.666 (\color{red}$\uparrow$\color{black}) & 0.505 (\color{red}$\uparrow$\color{black}) & 0.634 (\color{red}$\uparrow$\color{black}) \\
    Aligned-MTL \cite{alignedmtl} & 0.400 (\color{red}$\uparrow$\color{black}) & 0.547 (\color{red}$\uparrow$\color{black}) & 0.477 (\color{blue}$\downarrow$\color{black}) & 0.610 (\color{blue}$\downarrow$\color{black}) & 0.460 (\color{blue}$\downarrow$\color{black}) & 0.580 (\color{blue}$\downarrow$\color{black}) \\
    IMTL \cite{imtl} & 0.410 (\color{red}$\uparrow$\color{black}) & 0.534 (\color{red}$\uparrow$\color{black}) & \textbf{0.550} (\color{red}$\uparrow$\color{black}) & 0.667 (\color{red}$\uparrow$\color{black}) & \textbf{0.541} (\color{red}$\uparrow$\color{black}) & \textbf{0.658} (\color{red}$\uparrow$\color{black}) \\
    \hline
    RLW \cite{randomweighting} & 0.360 (\color{blue}$\downarrow$\color{black}) & 0.504 (\color{blue}$\downarrow$\color{black}) & 0.499 (\color{blue}$\downarrow$\color{black}) & 0.649 (--) & 0.466 (\color{blue}$\downarrow$\color{black}) & 0.614 (--) \\
    DWA \cite{mtan} & 0.390 (\color{red}$\uparrow$\color{black}) & 0.533 (\color{red}$\uparrow$\color{black}) & \textbf{0.533} (\color{red}$\uparrow$\color{black}) & 0.664 (\color{red}$\uparrow$\color{black}) & 0.499 (\color{red}$\uparrow$\color{black}) & 0.628 (\color{red}$\uparrow$\color{black}) \\
    Uncertainty \cite{uncertainty} & 0.206 (\color{blue}$\downarrow$\color{black}) & 0.349 (\color{blue}$\downarrow$\color{black}) & 0.345 (\color{blue}$\downarrow$\color{black}) & 0.507 (\color{blue}$\downarrow$\color{black}) & 0.319 (\color{blue}$\downarrow$\color{black}) & 0.482 (\color{blue}$\downarrow$\color{black}) \\
    FAMO \cite{famo} & \textbf{0.431} (\color{red}$\uparrow$\color{black}) & \textbf{0.564} (\color{red}$\uparrow$\color{black}) & 0.517 (--) & 0.623 (\color{blue}$\downarrow$\color{black}) & 0.510 (\color{red}$\uparrow$\color{black}) & 0.613 (--) \\
    \hline
    CosReg \cite{cosreg} & 0.387 (--) & 0.545 (\color{red}$\uparrow$\color{black}) & 0.522 (--) & \textbf{0.672} (\color{red}$\uparrow$\color{black}) & 0.488 (--) & 0.631 (\color{red}$\uparrow$\color{black}) \\
    \hline
\end{tabular}

%% file: sections/3_benchmark/2_city_scapes_nyu_v2.tex
\subsection{Experiments on CityScapes and NYU-v2}

We largely base our experiment setup for these two datasets on existing work in the literature and refer the readers to Appendix \ref{suppl:setup_benchmarks} for details.
Full results on CityScapes and NYU-v2 are summarized in Tables \ref{table:benchmark_city_scapes} and \ref{table:benchmark_nyu_v2}, respectively.
Results have shown that CosReg \cite{cosreg} consistently reaches top 3 on both datasets and all evaluation metrics. Moreover, GradDrop \cite{graddrop} also achieved top 3 scores except normal estimation (NE) on NYU-v2 \cite{nyu_v2}, but still improved the baseline.

\begin{table}
    \centering
    \small{\input{tables/benchmark_city_scapes}}
    \caption
    {
        Benchmark results with ResNet18 backbone on CityScapes \cite{cityscapes}. DE stands for depth estimation, evaluated by $L_{1}$ distance; SS stands for semantic segmentation, evaluated by mIoU; and IS stands for instance segmentation, evaluated by $L_{1}$ distance.
    }
    \label{table:benchmark_city_scapes}
\end{table}

\begin{table}
    \centering
    \small{\input{tables/benchmark_nyu_v2}}
    \caption
    {
        Benchmark results with ResNet18 backbone on NYU-v2 \cite{nyu_v2}. DE stands for depth estimation, evaluated by $L_{1}$ distance; NE stands for normal estimation, evaluated by angle in degrees, and SS stands for semantic segmentation, evaluated by mIoU.
    }
    \label{table:benchmark_nyu_v2}
\end{table}

%% file: tables/benchmark_city_scapes.tex
\begin{tabular}{c|P{1.4cm}P{1.4cm}P{1.4cm}}
    \hline
     & DE ($\downarrow$) & SS ($\uparrow$) & IS ($\downarrow$) \\
    \hline
    Baseline & 1.453 & 15.0\% & 0.131 \\
    \hline
    RGW \cite{randomweighting} & 1.465 & 14.9\% & 0.131 \\
    PCGrad \cite{pcgrad} & 1.462 & 15.0\% & 0.133 \\
    GradVac \cite{gradvac} & 1.466 & 15.0\% & 0.132 \\
    MGDA \cite{mgda} & 1.490 & 15.1\% & 0.134 \\
    CAGrad \cite{cagrad} & \textbf{1.446} & \textbf{15.3\%} & 0.128 \\
    GradDrop \cite{graddrop} & \textbf{1.367} & \textbf{15.5\%} & \textbf{0.117} \\
    Nash-MTL \cite{nashmtl} & 1.472 & 15.1\% & 0.132 \\
    Aligned-MTL \cite{alignedmtl} & 1.499 & 14.8\% & 0.134 \\
    IMTL \cite{imtl} & 1.467 & 15.0\% & 0.128 \\
    \hline
    RLW \cite{randomweighting} & 1.507 & 14.6\% & 0.133 \\
    DWA \cite{mtan} & 1.457 & 15.0\% & 0.131 \\
    Uncertainty \cite{uncertainty} & 15.597 & 1.4\% & 0.158 \\
    GradNorm \cite{gradnorm} & 1.454 & 15.0\% & 0.133 \\
    FAMO \cite{famo} & 15.059 & 1.3\% & \textbf{0.055} \\
    \hline
    CosReg \cite{cosreg} & \textbf{1.344} & \textbf{16.6\%} & \textbf{0.081} \\
    \hline
\end{tabular}

%% file: tables/benchmark_nyu_v2.tex
\begin{tabular}{c|P{1.4cm}P{1.4cm}P{1.4cm}}
    \hline
     & DE ($\downarrow$) & NE ($\downarrow$) & SS ($\uparrow$) \\
    \hline
    Baseline & 1.116 & 44.227 & 2.7\% \\
    \hline
    RGW \cite{randomweighting} & 1.111 & 43.759 & 2.7\% \\
    PCGrad \cite{pcgrad} & 1.125 & 43.834 & 2.7\% \\
    GradVac \cite{gradvac} & 1.118 & 43.954 & \textbf{2.8\%} \\
    MGDA \cite{mgda} & 1.155 & 44.632 & 2.6\% \\
    CAGrad \cite{cagrad} & 1.143 & 44.042 & 2.7\% \\
    GradDrop \cite{graddrop} & \textbf{1.093} & 43.912 & \textbf{2.9\%} \\
    Nash-MTL \cite{nashmtl} & 1.119 & 44.013 & 2.7\% \\
    Aligned-MTL \cite{alignedmtl} & 1.149 & 46.168 & 2.7\% \\
    IMTL \cite{imtl} & 1.120 & 44.040 & 2.6\% \\
    \hline
    RLW \cite{randomweighting} & 1.541 & \textbf{43.524} & 2.4\% \\
    DWA \cite{mtan} & 1.120 & 44.196 & 2.6\% \\
    Uncertainty \cite{uncertainty} & 9.099 & 47.985 & 0.3\% \\
    GradNorm \cite{gradnorm} & 1.115 & 45.786 & 2.6\% \\
    FAMO \cite{famo} & \textbf{1.097} & \textbf{39.148} & 0.5\% \\
    \hline
    CosReg \cite{cosreg} & \textbf{0.896} & \textbf{40.586} & \textbf{3.5\%} \\
    \hline
\end{tabular}

%% file: sections/4_surrogate.tex
\section{Generalizability of Fast Gradient Surrogate}
\label{sec:surrogate}

\begin{table*}
    \centering
    \small{\input{tables/surrogate}}
    \caption
    {
        Benchmark results with ResNet18 backbone on MetaGraspNet dataset but all gradients in the algorithms replaced with feature-level gradients. Table entries are relative performance change compared to their parameter-level gradient counterparts.
    }
    \label{table:surrogate}
\end{table*}

It has been a common technique to replace $\nabla_{\theta^{\text{sh}}}\mathcal{L}_{i}$ with $\nabla_{\mathcal{Z}}\mathcal{L}_{i}$ for reduced computation cost \cite{mgda, alignedmtl, rotograd}. While it might be reasonable to do so due to chain rule and sub-additivity of norms \cite{mgda, alignedmtl}, other methods \cite{nashmtl} has reported significant performance degrade with feature-level gradients in their method. To the best of our knowledge, there is no prior work addressing the generalizability of this approximation via comprehensive empirical study across a large basket of existing algorithms.

We compared the performance using $\nabla_{\theta^{\text{sh}}}\mathcal{L}_{i}$ versus $\nabla_{\mathcal{Z}}\mathcal{L}_{i}$ on 8 optimization algorithms on the MetaGraspNet \cite{metagraspnetv2} dataset. Full results are summarized in Table \ref{table:surrogate}. Results have displayed the following: Firstly, only MGDA \cite{mgda} and Aligned-MTL \cite{alignedmtl} achieved consistent performance gain under all evaluation metrics, and these are exactly the two selected methods that argued that feature-level gradients could be used as an upper bound (up to scaling) during the optimization process. On the other hand, GradVac \cite{gradvac}, RGW \cite{randomweighting}, and IMTL \cite{imtl} got significant performance degradation and hence this surrogate is clearly not applicable to these algorithms. We conclude that this fast gradient surrogate is not generalizable and we encourage more theoretical analysis to be done for each method.

%% file: tables/surrogate.tex
\begin{tabular}{c|cccccc}
    \hline
     & BBox mAP & BBox mAR & VMask mAP & VMask mAR & AMask mAP & AMask mAR \\
    \hline
    RGW \cite{randomweighting} & -6.4\% & -3.6\% & -6.9\% & -4.2\% & -5.9\% & -3.4\% \\
    PCGrad \cite{pcgrad} & -1.4\% & -0.7\% & 0.7\% & -0.2\% & 3.2\% & 3.0\% \\
    GradVac \cite{gradvac} & -13.9\% & -7.8\% & -11.2\% & -7.6\% & -14.2\% & -10.6\% \\
    MGDA \cite{mgda} & 14.3\% & 6.8\% & 17.8\% & 12.0\% & 19.8\% & 14.1\% \\
    CAGrad \cite{cagrad} & -0.9\% & -2.9\% & 2.9\% & 2.6\% & 5.1\% & 4.8\% \\
    Aligned-MTL \cite{alignedmtl} & 2.5\% & 1.3\% & 8.1\% & 5.9\% & 8.3\% & 7.7\% \\
    IMTL \cite{imtl} & -7.4\% & -1.5\% & -8.2\% & -1.6\% & -8.1\% & -1.1\% \\
    \hline
    CosReg \cite{cosreg} & 0.2\% & -2.3\% & 1.5\% & 0.3\% & 0.8\% & -0.4\% \\
    \hline
\end{tabular}

%% file: sections/5_method.tex
\section{Feature Disentanglement Measure}\label{sec:method}

In this section, we propose a novel measurement using feature disentanglement for identifying the challenges in MTL problems. To the best of our knowledge, we are the first to explicitly quantify the severity of feature disentanglement and to monitor its training dynamics.

The intuition behind lies in the gap between Single-task Learning (STL) and MTL. STL is easier in the sense that each task could be solved independently. The gap between MTL and STL could be bridged by allocating disjoint subsets of the extracted features for each task, while still using a single backbone.
We propose to understand the challenges in MTL from the perspective of learnt shared representation and study MTL by asking fundamental question: \textit{what kind of features are beneficial for a given set of tasks?}

\begin{table*}
    \centering
    {\small\input{tables/ranking_metagraspnet}}
    \caption
    {
        Ranking similarity results on MetaGraspNet \cite{metagraspnetv2} dataset.
    }
    \label{table:ranking_metagraspnet}
\end{table*}

\input{sections/5_method/1_preliminaries}
\input{sections/5_method/2_feature_disentanglement}
\input{sections/5_method/3_ranking_similarity}
\input{sections/5_method/4_qualitative}
\input{sections/5_method/5_quantitative}

%% file: tables/ranking_metagraspnet.tex
\begin{tabular}{c|cccccc}
    \hline
     & BBox mAP & BBox mAR & VMask mAP & VMask mAR & AMask mAP & AMask mAR \\
    \hline
    GDS & 64.3\% & 63.3\% & 50.5\% & 60.0\% & 56.7\% & 51.0\% \\
    \hline
    GMS & \textbf{67.6\%} & \textbf{70.5\%} & 53.8\% & 59.5\% & 50.5\% & 59.0\% \\
    \hline
    FD & 56.7\% & 58.6\% & \textbf{65.7\%} & \textbf{65.7\%} & \textbf{65.2\%} & \textbf{69.0\%} \\
    \hline
\end{tabular}

%% file: sections/5_method/1_preliminaries.tex
\subsection{Preliminaries}\label{subsec:preliminaries}

The literature is familiar with how task conflicts and task dominance are defined for two tasks. In this section, we make it clear how these measures are defined for a set of $T$ tasks.

\noindent\textbf{Task Conflicts}.
We follow \cite{pcgrad, gradvac, cosreg} and define the Gradient Direction Similarity (GDS) measure for $T$ tasks as
\begin{subequations}
    \begin{align}
        \alpha_{ij} & := \frac
        {\langle
            \nabla_{\theta^{\text{sh}}}\mathcal{L}_{i},
            \nabla_{\theta^{\text{sh}}}\mathcal{L}_{j}
        \rangle}
        {
            \|\nabla_{\theta^{\text{sh}}}\mathcal{L}_{i}\|_{2}
            \|\nabla_{\theta^{\text{sh}}}\mathcal{L}_{j}\|_{2}
        }
        \text{ for } i, j \in [T]; \\
        \text{GDS} & := \frac{1}{T(T-1)}\sum\bigg\{\alpha_{ij}: i, j \in [T], i \neq j\bigg\},
    \end{align}
    \label{eq:tc}
\end{subequations}
where $\alpha_{ij} \in [-1, +1]$ is the cosine value of the angle between $\nabla_{\theta^{\text{sh}}}\mathcal{L}_{i}$ and $\nabla_{\theta^{\text{sh}}}\mathcal{L}_{j}$ and quantifies the relationship between the \textit{directions} of the task gradients. A lower GDS score indicates less agreement between the supervision from different losses.

\noindent\textbf{Task Dominance}.
We follow \cite{pcgrad} and define the Gradient Magnitude Similarity (GMS) measure for $T$ tasks as
\begin{subequations}
    \begin{align}
        \beta_{ij} & := \frac{2
            \|\nabla_{\theta^{\text{sh}}}\mathcal{L}_{i}\|_{2}
            \|\nabla_{\theta^{\text{sh}}}\mathcal{L}_{j}\|_{2}
        }
        {
            \|\nabla_{\theta^{\text{sh}}}\mathcal{L}_{j}\|_{2}^{2} + 
            \|\nabla_{\theta^{\text{sh}}}\mathcal{L}_{j}\|_{2}^{2}
        },
        \text{ for } i, j \in [T]; \\
        \text{GMS} & := \frac{1}{T(T-1)}\sum\bigg\{\beta_{ij}: i, j \in [T], i \neq j\bigg\}
    \end{align}
    \label{eq:td}
\end{subequations}
where $\beta_{ij} \in [0, 1]$ quantifies the relationship between the \textit{magnitudes} of the task gradients. A lower GMS score indicates less aligned learning pace between the losses.

%% file: sections/5_method/2_feature_disentanglement.tex
\subsection{Method: Feature Disentanglement Measure}
\label{subsec:feature_disentanglement}

Our method, Feature Disentanglement measure, is defined as follows.
Given an extracted feature $\mathcal{Z}$ and a task loss $\mathcal{L}_{i}$, how much a change in $\mathcal{Z}_{j}$ can affect $\mathcal{L}_{i}$ can be measured by $|\nabla_{\mathcal{Z}_{j}}\mathcal{L}_{i}| \in \mathbb{R}_{\geq0}$, and the tensor $|\nabla_{\mathcal{Z}}\mathcal{L}_{i}|$ of gradient magnitudes, which has the same shape as $\mathcal{Z}$, can be interpreted as a saliency map on $\mathcal{Z}$, as inspired by GradCam \cite{gradcam} in the Explainable AI (XAI) literature.
At location $j$, we can quantify the entropy of the saliencies across $T$ tasks by
\begin{subequations}\label{eq:entropy0}
    \begin{align}
        \mathcal{E}_{j}(\mathcal{Z}) & := -\sum_{i=1}^{T}p_{ij}\log p_{ij}, \text{ where } \\
        p_{ij} & := |\nabla_{\mathcal{Z}_{j}}\mathcal{L}_{i}| / \sum_{k=1}^{T}|\nabla_{\mathcal{Z}_{j}}\mathcal{L}_{k}|
    \end{align}
\end{subequations}
The feature disentanglement measure for the entire shared representation $\mathcal{Z}$ is defined to be the average entropy across all positions:
\begin{equation}\label{eq:entropy}
    \text{FD} := \texttt{mean}\bigg\{\mathcal{E}_{j}(\mathcal{Z}): j \in [\dim(\mathcal{Z})]\bigg\},
\end{equation}
where $\dim(\mathcal{Z})$ denotes the dimension of the Euclidean space $\mathcal{Z}$ lies in.
A lower feature disentanglement measurement indicates that activations are salient to fewer tasks, and hence larger disentangled-ness.
Note that monitoring task conflicts and task dominance is extremely expensive as they require back propagation through until the first layer $T$ times to compute $\nabla_{\theta^{\text{sh}}}\mathcal{L}_{i}$. However, the feature disentanglement measure (ours) is a lot cheaper as it only replies on $\nabla_{\mathcal{Z}}\mathcal{L}_{i}$ and only need to back propagates through the $T$ prediction heads.
An illustration of this definition is shown in Figure \ref{fig:def_fd}.

%% file: sections/5_method/3_ranking_similarity.tex
\subsection{Evaluation Protocol: Ranking Similarity}
\label{subsec:ranking_similarity}

In order to quantitatively evaluate the faithfulness of different measures (GDS, GMS, and FD) for revealing the challenges in MTL problems, we propose Ranking Similarity to quantify the alignment against test-time performance.

\noindent\textbf{Definition}. Given a set of $n$ scalars $A := \{a_{1}, ..., a_{n}\} \subset \mathbb{R}$ and two rankings $R_{1}, R_{2}: A \to [n]$ on $A$, we define the \textit{ranking similarity} $\mathcal{S}(R_{1}, R_{2})$ between $R_{1}$ and $R_{2}$ to be the following average:
\begin{equation}
\begin{aligned}
    & \mathcal{S}(R_{1}, R_{2}) := \\
    & \frac{1}{n(n-1)}\sum_{\substack{i,j=1\\i\neq j}}^{n}\mathbb{I}\bigg[R_{1} \text{ and } R_{2} \text{ agree on } a_{i} \text{ and } a_{j}\bigg],
\end{aligned}
\end{equation}
where for any $i, j \in [n]$ with $i \neq j$, $R_{1}$ and $R_{2}$ agree on $a_{i}$ and $a_{j}$ if and only if ``$R_{1}(a_{i}) > R_{1}(a_{j})$'' and ``$R_{2}(a_{i}) > R_{2}(a_{j})$'' have the same truth value. i.e., $\mathcal{S}(R_{1}, R_{2})$ is the percentage of pairs $(a_{i}, a_{j})$ with the same ordering under $R_{1}$ and $R_{2}$.
Larger ranking similarity means more agreement under different ranking methods.

\noindent\textbf{Symmetry}. Let $R_{1}$ and $R_{2}$ be two arbitrary rankings and let $R_{2}'$ be the reverse of $R_{2}$. Then $R_{1}$ and $R_{2}$ agree on $(a_{i}, a_{j})$ if and only if $R_{1}$ and $R_{2}'$ disagree. So $\mathcal{S}(R_{1}, R_{2}') = 1-\mathcal{S}(R_{1}, R_{2})$. As we can always reverse a ranking when making comparisons, we only consider $|\mathcal{S}(R_{1}, R_{2}) - 0.5|$. The larger the absolute value, the clearer $R_{1}$ and $R_{2}$ aligns. We only report this symmetric version between test-time performance scores and MTL challenge measures.

%% file: sections/5_method/4_qualitative.tex
\subsection{Qualitative Results}\label{subsec:qualitative}

\begin{figure}[t]
    \centering
    \includegraphics[width=\linewidth]{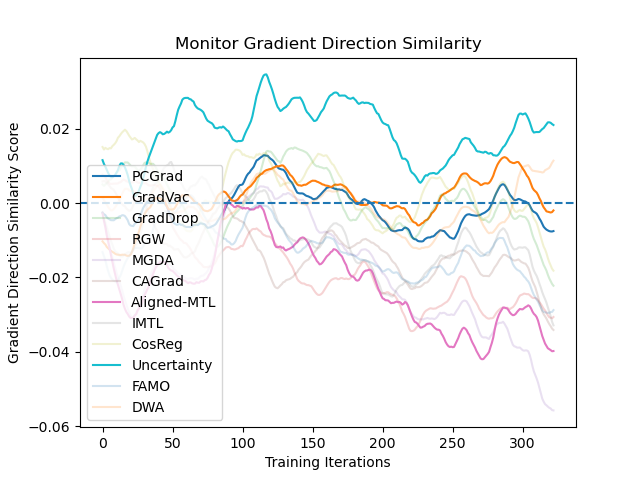}
    \caption{Training dynamics of GDS using gradients w.r.t. shared parameters.}
    \label{fig:GDS_analysis}
\end{figure}


\begin{figure}[t]
    \centering
    \includegraphics[width=\linewidth]{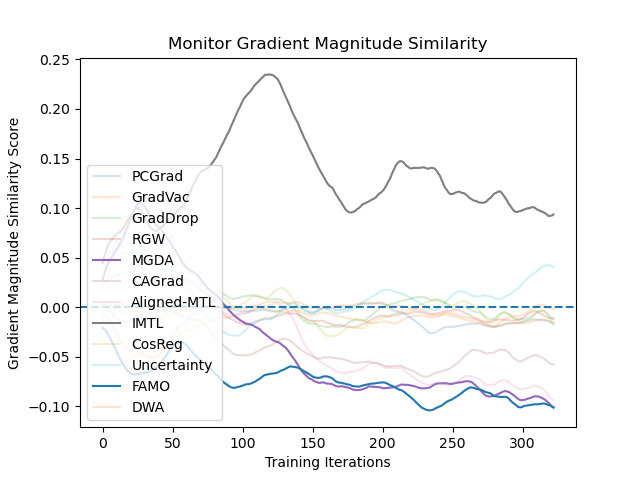}
    \caption{Training dynamics of GMS using gradients w.r.t. shared parameters.}
    \label{fig:GMS_analysis}
\end{figure}


\begin{figure}[t]
    \centering
    \includegraphics[width=\linewidth]{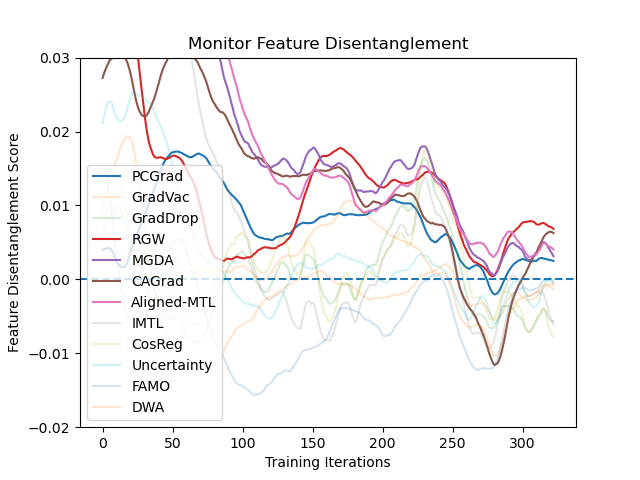}
    \caption{Training dynamics of Feature Disentanglement (FD) using gradients w.r.t. shared parameters.}
    \label{fig:fd_analysis}
\end{figure}


We show interesting examples and qualitatively demonstrate the effectiveness of our method from the training dynamics on MetaGraspNet \cite{metagraspnetv2} dataset. Trajectories are plotted in Figures \ref{fig:GDS_analysis}, \ref{fig:GMS_analysis}, and \ref{fig:fd_analysis}. For clarity, we only plot the relative values to the baseline method. All curves are smoothened by taking the moving average with window size equal to $1/10$ of the total trajectory length. We also added transparency to those not of interest and emphasized ones from which we make important qualitative observations. See Appendix \ref{suppl:setup_dynamics} for more details on how trajectories were plotted.

\noindent\textbf{Task Conflicts and Task Dominance}.
Figure \ref{fig:GDS_analysis} and Table \ref{table:benchmark_metagraspnet} have shown that Uncertainty Weighting, which had poor performance on the test set (Table \ref{table:benchmark_metagraspnet}), and GradVac, which had stronger performance, both achieved high GDS scores. In contrast, PCGrad and Aligned-MTL, which are hard to conclude one is superior to the other on the test set, lie far apart in the GDS plot.
Figure \ref{fig:GMS_analysis} and Table \ref{table:benchmark_metagraspnet} have shown that MGDA and FAMO achieved almost the same GDS curves, but FAMO achieved significantly better performance results than MGDA.

\noindent\textbf{Feature Disentanglement}. Figure \ref{fig:fd_analysis} and Table \ref{table:benchmark_metagraspnet} have show that most methods applied parameter-level gradients achieved feature entangled-ness lower than baseline close to the end of training, with the exception for RGW, PCGrad, MGDA, and Aligned-MTL These four methods form exactly the complement of the three methods that achieved performance gain among the gradient manipulation methods, as reported in Table \ref{table:benchmark_metagraspnet}. Nevertheless, clear decrease trends are displayed in PCGrad, MGDA, and Aligned-MTL.
This provides strong evidence that the previous success in these methods can be attributed to learning disentangled features for down stream tasks.

%% file: sections/5_method/5_quantitative.tex
\subsection{Quantitative Results}\label{subsec:quantitative}

\begin{table}
    \centering
    {\small\input{tables/ranking_city_scapes}}
    \caption
    {
        Ranking similarity results on CityScapes \cite{cityscapes}.
    }
    \label{table:ranking_city_scapes}
\end{table}

\begin{table}
    \centering
    {\small\input{tables/ranking_nyu_v2}}
    \caption
    {
        Ranking similarity results on NYU-v2 \cite{nyu_v2}.
    }
    \label{table:ranking_nyu_v2}
\end{table}

When defining the ordering of training trajectories, we took the mean of the last 50 elements, which is the closest to the end of training.
With RS, we are able to quantitatively report the faithfulness of the MTL measures. Results on three datasets are summarized in Tables \ref{table:ranking_metagraspnet}, \ref{table:ranking_city_scapes}, and \ref{table:ranking_nyu_v2}. Results have shown that on MetaGraspNet \cite{metagraspnetv2} dataset, FD has lower ranking similarities for bounding box predictions compared to GDS or GMS, but consistently out-performs traditional GDS and GMS for visible and amodal mask predictions). On CityScapes \cite{cityscapes}, FD out-performed GMS and GDS on semantic segmentation and instance segmentation, and out-performed GMS on depth estimation. On NYU-v2, FD still out-performed GMS on depth estimation and normal estimation.

%% file: tables/ranking_city_scapes.tex
\begin{tabular}{c|P{1.4cm}P{1.4cm}P{1.4cm}}
    \hline
     & DE & SS & IS \\
    \hline
    GMS & 56.8\% & 58.2\% & 61.3\% \\
    GDS & \textbf{58.2\%} & 55.3\% & 55.7\% \\
    FD & 57.3\% & \textbf{59.0\%} & \textbf{61.3\%} \\
    \hline
\end{tabular}

%% file: tables/ranking_nyu_v2.tex
\begin{tabular}{c|P{1.4cm}P{1.4cm}P{1.4cm}}
    \hline
     & DE & NE & SS \\
    \hline
    GMS & 61.3\% & 57.5\% & \textbf{67.8\%} \\
    GDS & \textbf{70.0\%} & \textbf{72.7\%} & 60.5\% \\
    FD & 67.9\% & 70.6\% & 57.9\% \\
    \hline
\end{tabular}

%% file: sections/6_conclusions.tex
\section{Conclusions}\label{sec:conclusions}

In this paper, we first conducted large-scale experiments with ResNet18 \cite{resnet} backbone on CityScapes \cite{cityscapes}, NYU-v2 \cite{nyu_v2}, a new large-sized and complex test ground, MetaGraspNet \cite{metagraspnetv2} dataset. We showed that GradDrop \cite{graddrop} and CosReg \cite{cosreg} were the best-performing methods on all three datasets.
Secondly, we compared each method with parameter-level gradients versus feature-level gradients and showed that only MGDA \cite{mgda} and Aligned-MTL \cite{alignedmtl} which have theoretical guarantees achieved a performance gain, while others were either similar or had significant performance degrade.
Lastly, we proposed a novel efficient method to identify the challenges in MTL, and showed its faithfulness for visible and amodal mask predictions on MetaGraspNet and for semantic and instance segmentation on CityScapes. We leave proposing improved measures based on shared representations as a future direction.

%% file: sections/suppl.tex
\clearpage
\setcounter{page}{1}
\maketitlesupplementary

\section{Experiment Setup for Benchmark Results}
\label{suppl:setup_benchmarks}

All experiments were done on a pool of GPUs including NVIDIA RTX 6000 Ada Generation, NVIDIA RTX A6000, and NVIDIA GeForce RTX 4090.
Throughout, we fixed the backbone to be ResNet18 \cite{resnet} with weights pretrained on ImageNet \cite{imagenet}, and fixed the optimizer to be the PyTorch SGD optimizer with learning rate $1.0 \times 10^{-4}$ and momentum $0.9$. Linear warmup learning rate scheduler was also applied in all experiments. All images, including training, validation, and testing, are resized to $512 \times 512$. We used batch size of 4 for experiments on MetaGraspNet \cite{metagraspnetv2} and batch size of 32 for experiments on CityScapes \cite{cityscapes} and NYU-v2 \cite{nyu_v2}.

\section{Experiment Setup for Training Dynamics of MTL Measures}
\label{suppl:setup_dynamics}

To save computation, we only compute these measures every 10 training iterations.
With around $3k$ training iterations on the MetaGraspNet \cite{metagraspnetv2} dataset, we get around $300$ data points on the trajectories.